\documentclass[lettersize,journal]{IEEEtran}
\usepackage{amsmath,amsfonts}
\usepackage{algorithm}
\usepackage{algorithmicx}
\usepackage{bm} 
\usepackage{array}
\usepackage[caption=false,font=normalsize]{subfig}
\usepackage{textcomp}
\usepackage{stfloats}
\usepackage[square,numbers]{natbib}
\usepackage{url}
\usepackage{authblk}
\usepackage{booktabs}
\usepackage{verbatim}

\usepackage{amssymb}
\usepackage{cite}
\usepackage{algpseudocode}
\usepackage[pagebackref,breaklinks,colorlinks]{hyperref}
\usepackage{multicol} 
\usepackage{multirow}
\usepackage{diagbox}
\usepackage{graphicx}
\usepackage{caption}
\usepackage{diagbox}
\usepackage[table]{xcolor} 
\usepackage{xcolor}

\definecolor{lightpurple}{RGB}{229,204,255} 
\allowbreak

\begin{document}

\title{Local Background Features Matter in Out-of-Distribution Detection}

\author[1,3,4$\dagger$]{Jinlun Ye}
\author[1,3,4$\dagger$]{Zhuohao Sun}
\author[2$\ddagger$]{Yiqiao Qiu\thanks{$\dagger$ Authors contributed equally.}\thanks{$\ddagger$ Works not related to role at Amazon.}}
\author[5]{Qiu Li}
\author[1]{Zhijun Tan}
\author[1,3,4*]{Ruixuan Wang\thanks{* Corresponding author: wangruix5@mail.sysu.edu.cn}}

\affil[1]{School of Computer Science and Engineering, Sun Yat-sen University, Guangzhou, Guangdong, 510275, China}
\affil[2]{Amazon Web Services, Cupertino, CA 95014, United States of America}
\affil[3]{Peng Cheng Laboratory, Shenzhen, Guangdong, 518066, China}
\affil[4]{Key Laboratory of Machine Intelligence and Advanced Computing, MOE, Guangzhou, Guangdong, 510275, China}
\affil[5]{China United Network Communications Corporation Limited Guangdong Branch, Guangzhou, Guangdong, 510000, China}

\maketitle
\begin{abstract} 
Out-of-distribution (OOD) detection is crucial when deploying deep neural networks in the real world to ensure the reliability and safety of their applications.
One main challenge in OOD detection is that neural network models often produce overconfident predictions on OOD data. 
While some methods using auxiliary OOD datasets or generating fake OOD images have shown promising OOD detection performance, they are limited by the high costs of data collection and training.
In this study, we propose a novel and effective OOD detection method that utilizes local background features as fake OOD features for model training. 
Inspired by the observation that OOD images generally share similar background regions with ID images, the background features are extracted from ID images as simulated OOD visual representations during training based on the local invariance of convolution.
Through being optimized to reduce the $L_2$-norm of these background features, the neural networks are able to alleviate the overconfidence issue on OOD data. 
Extensive experiments on multiple standard OOD detection benchmarks confirm the effectiveness of our method and its wide combinatorial compatibility with existing post-hoc methods, with new state-of-the-art performance achieved from our method.
\end{abstract}

\section{Introduction}
\label{sec:intro}

\IEEEPARstart{I}{n} recent years, deep learning has made extraordinary achievements in various fields including image recognition~\citep{AlexNet, ResNet, densenet, qiu2023sats}, autonomous driving~\citep{autonomous, chen2024E2EAutonomousDriving, bai2024AnythingInAnyScene}, and medical diagnosis~\citep{medical, xie2024TaskBN}. However, deep neural networks often encounter out-of-distribution (OOD) data that deviate from training data distributions~\citep{mos, Skeleton-OOD, LUST2022335} when deployed to the real world. These OOD data may compromise the stability of the model with potentially severe consequences, which reminds us of the importance of OOD detection for deep neural networks.

Many studies~\citep{ODIN, Optimal, ssod, PALM} of OOD detection have been conducted.
Most approaches train a network on ID data and design a scoring function for the pre-trained network model to detect whether any new data is OOD or is not in the test stage.
Some other works attempt to enhance OOD detection capabilities by incorporating regularization during model training. For example, CIDER~\citep{cider} leverages contrastive learning to construct more compact ID representations in the representation space, thereby improving OOD detection performance. However, neural networks trained solely on ID data typically produce overconfident predictions on OOD data.
To address this issue, some methods~\citep{OE, oe1, oe2} explore using auxiliary OOD datasets during training to provide explicit knowledge of OOD data. 
Due to the difficulty in acquiring real auxiliary OOD data in practical applications, some approaches such as TagFog~\citep{cjk1} and Dream-OOD~\citep{dreamood} attempt to generate fake OOD images for model training. However, the incorporation of fake OOD images still introduces additional computational overhead during training. 
Consequently, methods like VOS~\citep{VOS} and NPOS~\citep{NPOS} instead construct fake OOD 
features in the feature space for model training, achieving competitive detection performance. 

\begin{figure*}[t]
\centering
\includegraphics[width=0.8\textwidth]{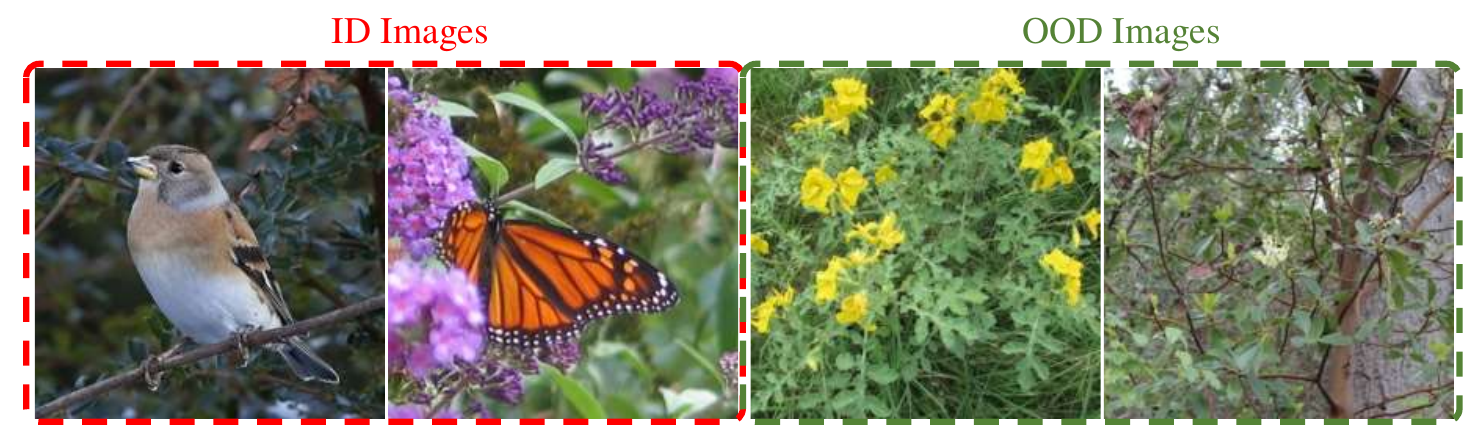} 
\caption{Examples of ID and OOD images. ID images and OOD images have similar background regions.}
\label{similar background images}
\end{figure*}

In this study, we firstly observe that the background regions of many OOD images are highly similar to those of ID images.
As shown in Figure~\ref{similar background images}, images of birds and butterflies often feature backgrounds of branches or flowers, similar to the OOD images depicting natural scenes.
If the neural network model focuses on these background areas irrelevant to ID categories during training, it may produce high feature activation for OOD data with similar backgrounds during the testing phase, leading to overconfident predictions on these OOD data. 
The visualized results in Figure~\ref{fig4:visualize} illustrate this phenomenon.
While recent studies~\citep{keyfeature, cjk2, EFOA} have treated that the background regions of ID images can be treated as fake OOD images for model training, this study leverages the local background features from ID data to simulate OOD feature representations, thereby enhancing OOD detection performance.
In this study, a fine-tuning strategy is proposed for a pre-trained neural network model that is initially trained on ID data using cross-entropy loss. 
During the fine-tuning stage, given the local invariance of the convolution operation, we first sample local background features from the feature maps output by the pre-trained model for ID images. 
Figure~\ref{t-sne} illustrates the distribution of ID features, OOD features, and local background features extracted from ID data at the penultimate layer (features space) based on the pre-trained model.
It's observed that the local background features extracted from ID images are similar to the features of real OOD images from the perspective of feature representation distribution, which means that they can be used as fake OOD features for model fine-tuning.  
Furthermore, we design a local feature fine-tuning loss to guide the model to reduce the activation of these background features, thereby encouraging the network to focus less on background information and ultimately mitigating the overconfidence issue on OOD data.
In summary, our main contributions are as follows:

\begin{figure}[t]
\centering
\includegraphics[width=1\linewidth, height=0.3\textheight]{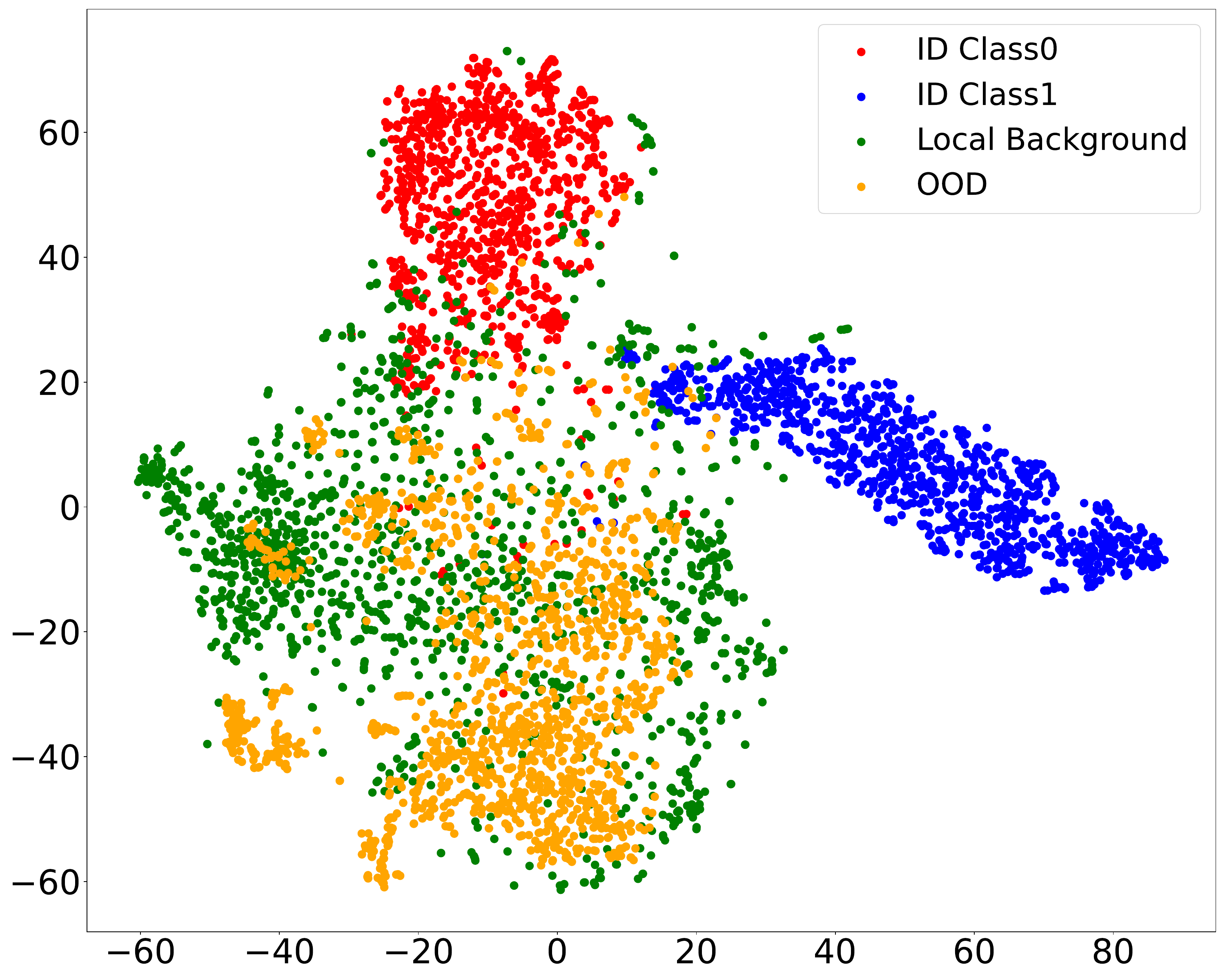} 
\caption{Visualization of the distribution of ID, OOD feature vectors, and local background feature vectors from ID data at the penultimate layer by t-SNE~\citep{tsne}. The model is ResNet-34~\citep{ResNet} trained on the CIFAR-10 dataset. The ID dataset is CIFAR-10 and OOD dataset is iSUN. Overall OOD features are far away from any ID class and similar to local background features.}
\label{t-sne}
\end{figure}

\section{Related Work}
OOD detection methods have evolved along two primary methodological directions.
One line of research is post-hoc approach which aims to address OOD detection by designing a scoring function for the output from the pre-trained network model, calculating an OOD score for each input sample~\citep{knn,vim,SHE,featurenorm, CORES}. For example, MSP~\citep{MSP} provides a simple baseline for OOD detection by using the maximum softmax probability output of the model. 
ODIN~\citep{ODIN} introduces two operations based on MSP called temperature scaling and input perturbation to separate OOD from ID samples.
Energy score~\citep{energy} uses the energy function of the logits for OOD detection. Mahalanobis score~\citep{mahala} uses the Mahalanobis distance between the feature vector of input data and the prototype feature vector of training data for OOD detection. KNN~\citep{knn} computes the $k$-th nearest neighbor distance between the feature vector of input sample and the feature vector of the training data. 
FeatureNorm~\citep{featurenorm} calculates the norm of the feature map to distinguish between ID and OOD data.

Another line of research performs OOD detection by training regularization~\citep{g-odin, learnconfidence, ssd, FlowCon, Greg}.
MoEP-AE~\citep{MoEP-AE} uses multiple mixtures of exponential power distributions to encode the features of ID classes and learn discriminative representations of ID classes.
SSD+~\citep{ssd} and CSI~\citep{csi} apply contrastive loss to train the model to get a more compact visual representation distribution for each ID class.
C-LMCL~\citep{C-LMCL} introduces a centralized large margin cosine loss which also constrains the feature representations to be more compact.
CIDER~\citep{cider} proposes a prototype-based representation learning framework for OOD detection that increases the distances between prototypes of different classes and encourages each sample to converge closer to the prototype of the class it belongs to, with superior performance achieved.

To alleviate the problem that the model trained solely on ID data has overconfidence in predicting OOD data as ID classes, LogitNorm~\citep{logitnorm} introduces to enforce a constant norm on the logit during model training.
Another straightforward approach~\citep{oe2, poem} leverages auxiliary OOD datasets for model training, which can help to build a stronger decision boundary between ID and OOD data.
However, these approaches often suffer from obtaining real auxiliary OOD datasets and struggle to guarantee the quality of auxiliary OOD datasets.
some methods attempt to generate fake OOD data using image generative models like Stable Diffusion~\citep{stablediffusion}, e.g., Dream-OOD~\citep{dreamood} and FodFom~\citep{cjk2} propose to generate fake OOD images that are semantically closer to ID images for model training.
VOS~\citep{VOS} and NPOS~\citep{NPOS} synthesize fake OOD features from boundary ID features in the feature space.
Furthermore, based on the observation that OOD data often cause abnormally high activation at the penultimate layer (feature vector space) of the network, ReAct~\citep{react} rectifies feature activation at an upper limit and reduces most of the activation values caused by OOD data. 
DICE~\citep{dice} leverage a sparsification strategy by clipping some noise units irrelevant to ID classes, resulting in improved separability in the OOD score distribution between ID and OOD data. 
LINe~\citep{LINe} employs the Shapley value~\citep{shapley} to measure each neuron’s contribution and reduces the effect of less important neurons in the feature vector and classifier head.
HIMPLoS~\citep{HIMPLoS} reduced feature activation of OOD data by masking features that are less important to the ID class, thereby widening the scoring gap between ID and OOD data.
These works attempt to leverage threshold clipping operations or sparsification strategies to minimize the feature activation of the model for OOD data.
Inspired by the research above, we introduce a simple yet effective framework for OOD detection that can extract fake OOD features from the local features of ID data and regularize their $L_2$-norm, therefore reducing the feature activation of real OOD data.

\section{Preliminary}
\subsection{Model Pre-training}
Consider a deep neural network image classification model trained on a dataset $\mathcal{D} = \{ (\bm{x}_i, y_i)\}^N_{i=1}$, where $\bm{x}_i$ is the $i$-th training image and $y_i \in \{1, 2, \ldots, K\}$ stands for the corresponding class label. 

Typically, an image classification model consists of an image encoder and a classifier head.
Given an input image $\bm{x}$, a feature map $\bm{z} = h(\bm{x}) \in \mathbb{R}^{C \times H \times W}$ is obtained by the image encoder $h(\cdot)$, where $C$ is the channel of the feature map, $H$ and $W$ denote the height and width of the feature map. Then, the global feature vector denoted by $\overline{\bm{z}} \in \mathbb{R}^{C}$ can be computed after global average pooling (GAP) of the feature map $\bm{z}$. Finally, with the classifier head $f(\cdot)$, we can get the output logit vector $f(\overline{\bm{z}})$ and the final prediction probability is computed as follows,
\begin{equation}
p(y=k | \bm{x}) = \frac{\exp(f_k(\overline{\bm{z}}))}{\sum\limits_{m=1}\limits^{K} \exp(f_m(\overline{\bm{z}}))},
\end{equation}
where $f_k(\overline{\bm{z}})$ is the logit element of the predicted class $k$ in the logit vector$f(\overline{\bm{z}})$. During pre-training stage, we use the common cross-entropy loss $\mathcal{L}_{CE}$ to optimize the image encoder and the classifier head,
\begin{equation}
\mathcal{L}_{CE} = - \frac{1}{N}\sum_{i=1}^{N}  \log(p(y=y_i | \bm{x}_i)) \,,
\end{equation}
where $N$ is the number of all ID training images.

\subsection{Out-of-distribution (OOD) Detection}
When deploying the neural network model in the real world, new data might stem from a particular unknown distribution that varies from the distribution of the training data, i.e. with unknown class object presented.
Such data are out-of-distribution (OOD) and should not be predicted as any of the in-distribution (ID) classes learned during model training. 
The task of OOD detection is to identify whether any new data is ID or OOD. 

OOD detection can be regarded as a binary classification problem. Specifically, a scoring function $S(\bm{x}; f)$ can be designed to estimate the degree of any new data $\bm{x}$ belonging to any of the ID classes, where the function $f$ denotes the neural network model whose output is used as the input to the scoring function. With the scoring function $S(\bm{x}; f)$, OOD detection can be simply formulated as a binary classifier $g (\bm{x}; f)$ as below,
\begin{equation}
g (\bm{x}; f)=
\begin{cases}
\text{ID} & \text{ if } S (\bm{x}; f) \ge \gamma  \\
\text{OOD} & \text{ if } S (\bm{x}; f) < \gamma 
\end{cases},
\end{equation}
where data with higher scores $S (\mathrm{x}; f)$ are classified as ID data and lower scores are classified as OOD data, and $\gamma$ is the threshold hyperparameter.

\section{Methodology}

\begin{figure}[t]
\centering
\includegraphics[width=0.48\textwidth]{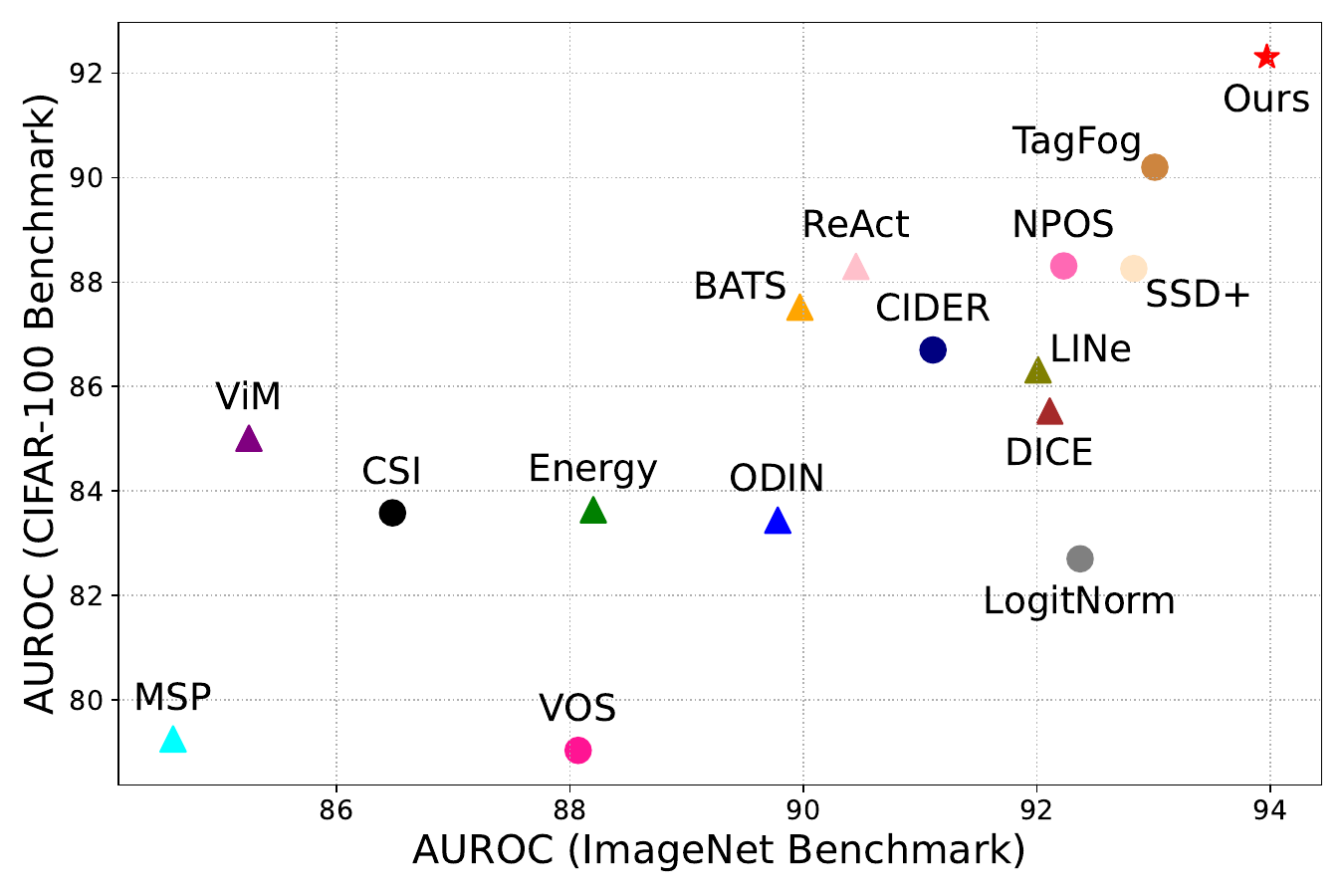} 
\caption{OOD detection performance from different methods on CIFAR-100 and ImageNet benchmarks. Larger AUROC means better performance. }
\label{performance}
\end{figure}

\begin{figure*}[t]
\centering
\includegraphics[width=1\textwidth]{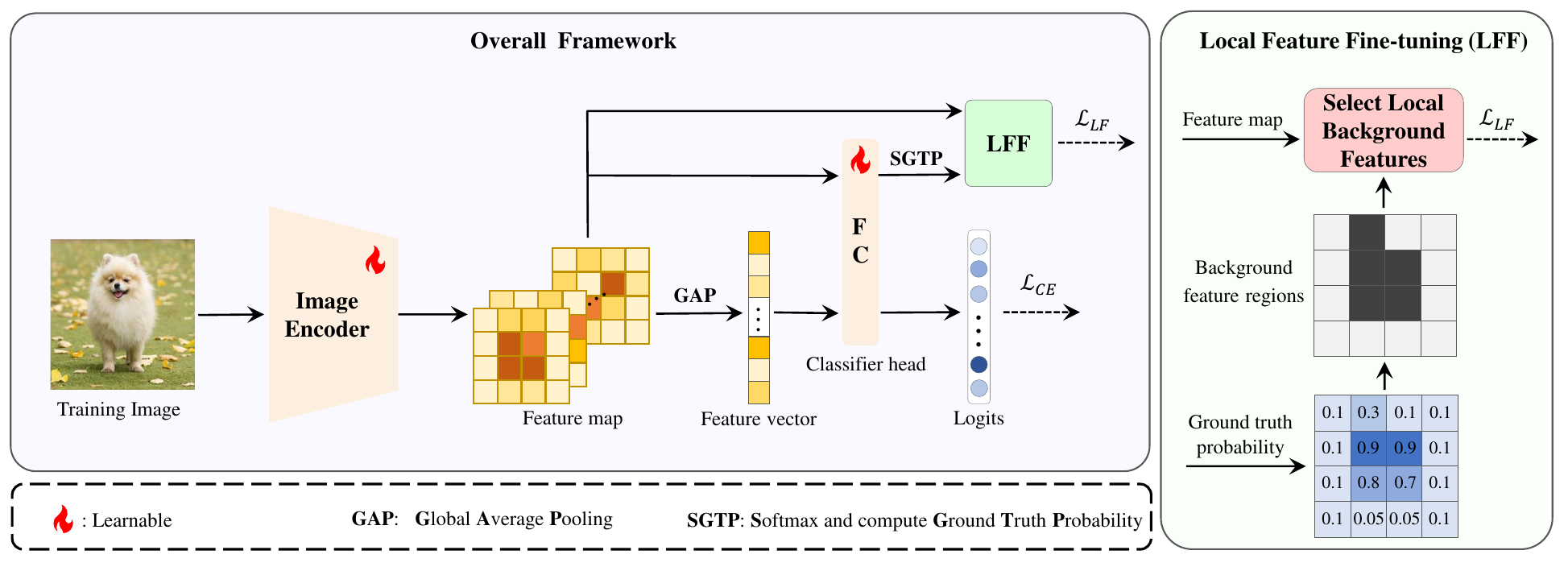} 
\caption{
Overview of the proposed framework. Given a pre-trained model and any ID image, we first select the local background feature vectors and obtain the global feature vector from the feature map. With the local background feature vectors and the logits computed by the global feature vector, the image encoder and classifier head are jointly trained by two loss: $\mathcal{L}_{CE}$, which guides the model to correct classification; $\mathcal{L}_{LFF}$, which constrains the $L_2$-norm of the local background feature vectors.
}
\label{method}
\end{figure*}

An overview of the proposed framework is illustrated in Figure~\ref{method}. 
After the classification model is well trained on ID dataset, it has already learned to pay attention to the foreground object region of the ID images therefore separating foreground class region (highly relevant to classification) and background region (less relevant or irrelevent to classification). Based on this circumstance, we introduce a fine-tuning strategy by jointly using the cross-entropy loss and the local background region feature fine-tuning loss to improve the OOD detection performance.
Specifically, in the fine-tuning stage, we firstly filter out the local background features from the feature maps of each training image.
As some studies~\citep{react, dice, LINe, HIMPLoS} reveal that due to the lack of OOD knowledge, OOD data trends to cause abnormally high activation in the feature space of the model, which is one of the main reasons for the model's overconfidence problem.
We utilize the local feature fine-tuning loss to restrict the $L_2$-norm of these background region features. In addition, to guarantee correct classification, we also incorporate the cross-entropy loss during model fine-tuning.
Figure~\ref{performance} summarizes the OOD detection performance of different methods on CIFAR-100 and ImageNet benchmarks, our method achieves state-of-the-art performance.

\subsection{Local Background Feature Extraction}
Given a pre-trained neural network model, we extract the local background features from images and employ them as fake OOD features during model fine-tuning.

Based on the local invariance of convolution, for a given input image $\bm{x}$, the feature map $\bm{z} \in \mathbb{R}^{C \times H \times W}$ output by the pre-trained model has a certain corresponding relationship with the features of the original image.
Therefore, the local features in the feature map corresponding to the original image can be divided into foreground features related to ID classes and background features denoting ID-irrelevant regions, and the background features can be extracted as fake OOD features.
The set of local feature vector indices for a feature map can be denoted as $J = \{1, 2, 3, \dots, H \times W \}$, where $H$ and $W$ denote the height and width of the feature map.
For an input image $\bm{x}$, the prediction probability of the model for each of its local feature vector $\bm{z}_j \in \mathbb{R}^{C} $ can be calculated as follows,
\begin{equation}
p_j(y=k | \bm{x}) = \frac{\exp(f_k(\bm{z}_j))}{\sum\limits_{m=1}\limits^{K} \exp(f_m(\bm{z}_j))},
\end{equation}
where $p_j(y=k | \bm{x})$ represents the probability that the model predicts the $j$-th local feature of the input image $\bm{x}$ to be of class $k$.

Normally, the pre-trained model can effectively identify the foreground features and classify them as the ground truth class with a relatively high probability. 
In contrast, since the local background features are irrelevant to any ID class, the prediction probability of the model for them as the ground truth class is particularly low.
Therefore, as shown in Figure~\ref{method} (right module), for each training input $\bm{x}_i$ in the training set, the local background feature vectors can be extracted through the following formula,
\begin{equation}
S = \{ \bm{z}_j | j \in J,  p_j(y=y_{i} | \bm{x}_i) < \delta\},
\end{equation}
where $y_{i}$ is the ground truth class corresponding to the training image $\bm{x}_i$ and $\delta$ is a hyperparameter representing the probability threshold.
Only the local features for which the probability of being predicted as the ground truth class by the model is less than the probability threshold $\delta$ are considered as local background features.
Finally, the background feature vectors of all the training images constitute the set
$S$.
\subsection{Local Feature Fine-Tuning}
The proposed local feature fine-tuning loss applies $L_2$-norm regularization to the local background feature vectors from the background feature vectors set $S$.
Specifically, the proposed local feature fine-tuning loss $\mathcal{L}_{LF}$ is designed as follows,
\begin{equation}
\mathcal{L}_{LF} = \frac{1}{|S|}\sum_{i=1}^{|S|} \max(||\bm{z}_i||_2-\mu, 0)
\end{equation}
where $|S|$ is the size of set $S$, $||\cdot||_2$ represents the $L_2$-norm operator of a n-dimensional vector, and $\mu$ is the margin hyperparameter of the $L_2$-norm.  
By minimizing $\mathcal{L}_{LF}$, we can minimize the $L_2$-norm of the background feature vectors below $\mu$, thereby reducing the activation of these background features.

\begin{figure*}[!t]
\centering
\includegraphics[width=1\textwidth]{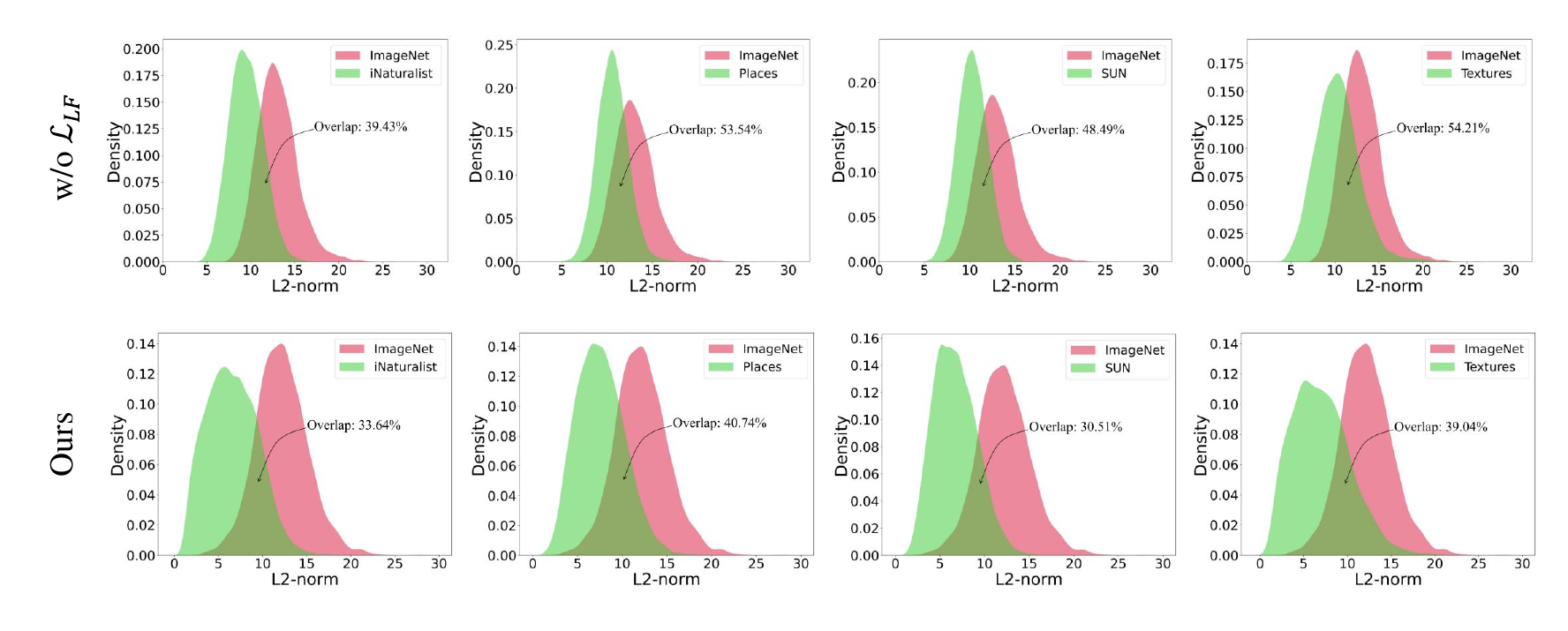} 
\caption{$L_2$-norm distributions of feature vectors for ID data (in red) or OOD data (in green) based on different models. ImageNet-100 is the ID dataset, and iNaturalist, SUN, Places, and Textures are four OOD datasets. w/o $\mathcal{L}_{LFF}$: model trained solely by $\mathcal{L}_{CE}$. Ours: model fine-tuned by $\mathcal{L}_{CE}$ and $\mathcal{L}_{LFF}$.}

\label{fig:l2norm}
\end{figure*}

Overall, the model will be trained by minimizing the joint loss function $\mathcal{L}$, 
\begin{equation}
\mathcal{L} = \mathcal{L}_{CE} + \lambda \mathcal{L}_{LF}
\end{equation}
where the coefficient hyperparameter $\lambda$ is used to balance the two loss terms.
Figure \ref{fig:l2norm} illustrates the $L_2$-norm distribution of the feature vectors output by the pre-trained model and the fine-tuned model for ID data and OOD data respectively.
Compared with the model (upper row in the figure) that was only pre-trained using the cross-entropy loss, the model (lower row in the figure) fine-tuned by the proposed loss significantly reduces the $L_2$-norm of the feature vectors output for real OOD data. 
This phenomenon indicates that the local background features extracted by our method can represent the output pattern of the model for OOD data, and when the model reduces the activation of these background features, it will also reduce the feature activation of these OOD images.
Thus, our method effectively alleviates the problem of the model's overconfident predictions for OOD data that possess the same background as ID data and improves the OOD detection performance.

\subsection{Scoring Function}
After fine-tuning, the model can be used for OOD detection by applying a scoring function.
While various scoring functions can be applied, the energy scoring function~\citep{energy} is used by default, i.e.,
\begin{equation}
    S(\bm{x}; f) = \log \sum_{m=1}^{K}\exp({f}_m (h(\bm{x}))) \,,
\label{eq:energy}
\end{equation}
where ${f}_m (h(\bm{x}))$ is the  $m$-th element in the logit vector ${f} (h(\bm{x}))$. 
In addition, since ReAct~\citep{react} achieves better performance by clipping the activation values of the global feature vector before calculating the energy score, we also apply the ReAct clipping operators during model inference.
Notably, our method is also compatible with other scoring functions such as MSP ~\citep{MSP} and ODIN ~\citep{ODIN}, which demonstrates the generalization ability of the proposed method.

\section{Experiments}

\begin{table*}[t]
\centering
\caption{Comparison between different methods on CIFAR-10 and CIFAR-100 benchmarks with ResNet-34. $\downarrow$ indicates smaller values mean better performance, $\uparrow$ indicates larger values mean better performance and $\dag$ indicates an additional fake OOD dataset is used. Post-hoc methods are shown in the white background and training regularization methods are shown in the gray background. Bold numbers are superior results and underlined numbers are the 2nd best results. All values are percentages.}
\label{tab:cifar-resnet benchmark}
\resizebox{\textwidth}{!}{%


\begin{tabular}{cccccccccccccccc}
\toprule
\multicolumn{1}{c}{\multirow{3}{*}{\begin{tabular}[c]{@{}c@{}}\textbf{ID Dataset}\end{tabular}}} & \multicolumn{1}{c}{\multirow{3}{*}{\textbf{Method}}} & \multicolumn{12}{c}{\textbf{OOD Datasets}} & \multicolumn{2}{c}{\multirow{2}{*}{\textbf{Average}}} \\ \cline{3-14} 
 \multicolumn{1}{c}{} & \multicolumn{1}{c}{} & \multicolumn{2}{c}{\textbf{SVHN}} & \multicolumn{2}{c}{\textbf{LSUN-C}} & \multicolumn{2}{c}{\textbf{LSUN-R}} & \multicolumn{2}{c}{\textbf{iSUN}} & \multicolumn{2}{c}{\textbf{Textures}} & \multicolumn{2}{c}{\textbf{Places365}} \\ 
 &  & FPR95$\downarrow$ & AUROC$\uparrow$ & FPR95$\downarrow$ & AUROC$\uparrow$ & FPR95$\downarrow$ & AUROC$\uparrow$ & FPR95$\downarrow$ & AUROC$\uparrow$ & FPR95$\downarrow$ & AUROC$\uparrow$ & FPR95$\downarrow$ & AUROC$\uparrow$ & FPR95$\downarrow$ & AUROC$\uparrow$ \\ \midrule
\multirow{19}{*}{\begin{tabular}[c]{@{}c@{}}CIFAR-10\end{tabular}} 
 & MSP & 33.79 & 94.18 & 20.47 & 96.86 & 42.62 & 92.31 & 45.17 & 91.71 & 49.95 & 89.66 & 53.69 & 87.84 & 40.95 & 92.09 \\
 & ODIN & 43.80 & 86.52 & 7.03 & 98.74 & 23.66 & 93.70& 26.03 & 93.18 & 45.73 & 83.19 & 52.08 & 82.55 & 33.06 & 89.64 \\
 & Mahalanobis & 41.38 & 94.10 & 92.99 & 88.48 & 52.67 & 92.54 & 52.26 & 92.41 & 38.95 & 94.24 & 55.89 & 90.19 & 55.69 & 91.99 \\
 & Energy & 20.65 & 95.25 & 5.32 & 99.05 & 25.64 & 94.13 & 28.17 & 93.46 & 38.83 & 88.91 & 41.49 & 88.26 & 26.69 & 93.17 \\
 & ViM & 29.66 & 95.06 & 49.00 & 93.84 & 38.13 & 94.00 & 37.49 & 93.84 & 28.19 & 94.94 & 47.58 & 90.78 & 38.35 & 93.75 \\
 & DICE & 36.67 & 90.64 & 6.67 & 98.70 & 35.13 & 92.79 & 40.95 & 90.92 & 50.73 & 86.60 & 49.65 & 84.70 & 36.63 & 90.72 \\ 
 & ReAct & 24.19 & 94.40 & 6.72 & 98.87 & 26.69 & 94.16 & 29.18 & 93.46 & 39.50 & 89.33 & 40.30 & 89.55 & 27.76 & 93.29 \\ 
 & DICE+ReAct & 38.22 & 90.64 & 6.67 & 98.70 & 34.83 & 92.79 & 40.95 & 89.92 & 50.73 & 87.17 & 49.67 & 84.72 & 36.85 & 90.72 \\
 & FeatureNorm & \underline{4.01} & 99.18 & \underline{0.53} & \textbf{99.87} & 44.30& 91.80 & 37.73 & 93.25 & 25.32 &  94.19& 66.69 & 80.28 & 29.76 & 93.10 \\
 & LINe & 27.64 & 94.98 & 3.41 & 99.31 & 54.03 & 88.20 & 56.53 & 87.04 & 54.26 & 86.95 & 61.93 & 80.02 & 42.97 & 89.42 \\
\cline{2-16}
\rowcolor{gray!20} \cellcolor{white!20} & CSI & 19.65 & 96.07 & 50.37 & 90.12 & 20.70 & 95.75 & 45.20 & 90.78 & 25.30 & 94.57 & 64.11 & 80.81 & 37.56 & 91.35 \\
\rowcolor{gray!20} \cellcolor{white!20} & SSD+ & \textbf{2.51}& \textbf{99.54} &46.72 &93.89& \underline{10.56} & \underline{97.83} & 28.34 & 95.87 & \textbf{9.27} & {\textbf{98.35}} & \underline{22.15}& \underline{95.37} & 19.93& 96.82 \\
\rowcolor{gray!20} \cellcolor{white!20} & VOS & 27.93 & 93.55 & 8.34 & 98.36 & 21.64 & 96.03 & 26.00 & 95.20 & 37.84 & 91.57 & 40.89 & 89.53 & 27.01 & 94.04 \\
\rowcolor{gray!20} \cellcolor{white!20} & LogitNorm & 17.40 & 96.96& \textbf{0.48} & {99.75} & {11.04} & {97.92}  & \underline{11.39} & \underline{97.86} & 31.51 & 94.71 & 29.90 & 94.31 & 16.95 & 96.93 \\
\rowcolor{gray!20} \cellcolor{white!20} & NPOS & 13.28 & 97.32 & 3.10 & 98.30& 17.76 & 96.38 & 15.44 & 96.99 & 26.64 & 95.74 & 35.77 & 93.21 & 18.67 & 96.32\\
\rowcolor{gray!20} \cellcolor{white!20} & CIDER & 4.93 & \underline{99.21} & 2.99 & 99.14 & 21.45 & 96.37 & 22.94 & 96.43 & 15.87 & {97.04} & 29.16 & 94.12 & 16.22 & 97.05 \\
\rowcolor{gray!20} \cellcolor{white!20} & TagFog$^\dag$ & 5.87 & 98.84& 9.05 & 98.31 & \textbf{2.02} & \textbf{99.44} & \textbf{8.87} & \textbf{98.30} & 16.63 & 96.88 & 24.56 & 94.52 & \textbf{11.17} &\textbf{97.72} \\
\rowcolor{gray!20} \cellcolor{white!20} & \textbf{Ours} & 7.71 & 98.59 & 1.14 & \underline{99.84} & 25.66 & 95.87 & 22.26 & 96.37 & \underline{13.34} & \underline{97.59} & \textbf{21.98} & \textbf{95.54} & \underline{15.35} & \underline{97.30} \\
\midrule

\multirow{19}{*}{\begin{tabular}[c]{@{}c@{}}CIFAR-100\end{tabular}} 
 & MSP & 81.31 & 77.61 & 76.52 & 80.65 & 74.39 & 82.08 & 75.99 & 80.96 & 81.72 & 76.90 & 79.81 & 77.30 & 78.29 & 79.25 \\
 & ODIN & 85.55 & 76.83 &  68.48 & 84.16 & 38.18 & 92.96 & 41.92 & 91.84 & 71.19 & 79.28 & 78.74 & 75.54 & 64.01 & 83.44 \\
 & Mahalanobis & 98.81 & 54.62 & 99.48 & 36.04 & 97.38 & 53.35 & 94.64 & 58.99 & 75.67 & 76.95 & 97.17 & 51.33 & 93.86 & 55.21 \\
 & Energy & 76.73 & 81.52 & 61.12 & 85.88 & 58.02 & 88.40 & 61.82 & 86.95 & 80.73 & 77.17 & 78.06 & 76.72 & 69.41 & 83.64 \\
 & ViM & 75.80 & 82.31 & 89.49 & 73.78 & 37.70 & 93.40 & 38.51 & 93.01 & 49.47 & 89.33 & 78.09 & 78.17 & 61.58 & 85.01 \\
 & DICE & 53.65 & 89.97 & 32.29 & 93.49 & 85.83 & 76.80 & 86.01 & 77.83 & 68.17 & 82.49 & 82.90 & 76.31 & 68.14 & 83.53 \\ 
 & ReAct & 43.10 & 92.22 & 52.61 & 87.32 & 41.60 & 91.58 & 41.95 & 91.29 & 53.31 & 87.65 & {70.80} & {79.73} & 50.56 & 88.30 \\ 
 & DICE+ReAct & 48.18 & 91.19 & 32.01 & 93.71 & 84.17 & 78.80 & 82.23 & 79.65 & 66.74 & 83.96 & 80.26 & 78.04 & 65.61 & 84.22 \\
 & FeatureNorm & 21.02 & 95.61 & \textbf{9.87} & \underline{98.21}& 97.06 & 67.20 & 91.79 & 73.87 & 45.23 & 84.79 & 92.64 & 61.02 & 59.60 & 80.12 \\
 & LINe & 39.97 & 91.17 & 26.51 & 94.02 & 61.02 & 86.90 & 62.62 & 86.16 & 55.18 & 86.80 & 81.81 & 72.90 & 54.52 & 86.32 \\
 \cline{2-16}
\rowcolor{gray!20} \cellcolor{white!20} & CSI & 44.53 & 92.65 & 86.12 & 77.34 & 75.58 & 83.78 & 76.62 & 84.98 & 61.61 & 86.47 & 79.08 & 76.27 & 70.59 & 83.58\\
\rowcolor{gray!20} \cellcolor{white!20} & SSD+& 20.92 & 96.42 & 75.06 & 86.01 & 37.95 & 93.55 & 80.97 & 83.73 & 54.24 & 90.23 & 78.75 & 79.64 & 57.98 & 88.26 \\
\rowcolor{gray!20} \cellcolor{white!20} & VOS & 83.52 & 81.24 & 79.40 & 80.39 & 77.81 & 78.50 & 78.34 & 78.23 & 84.35 & 77.81 & 79.77 & 78.01 & 80.53 & 79.03 \\
\rowcolor{gray!20} \cellcolor{white!20} & LogitNorm & 51.60 & 90.74 & \underline{10.57} & \textbf{98.22} & 89.78 & 77.67  & 84.39 & 77.55 & 81.56 & 74.45 & 80.30 & 77.59 & 66.37 & 82.70 \\
\rowcolor{gray!20} \cellcolor{white!20} & NPOS & \underline{18.56} & \underline{97.14} & 42.59 & 89.71& 48.37 & 89.33  & 47.61 & 88.52 & \textbf{{38.14}} & \textbf{{92.65}} & 79.75 & 72.52 & {45.84} & {88.31}\\
\rowcolor{gray!20} \cellcolor{white!20} & CIDER & 23.09 & 95.16 & 16.16 & 96.33 & 69.50 & 81.85 & 71.68 & 82.98 & \underline{43.87} & \underline{90.42} & 79.63 & 73.43 & 50.66 & 86.70 \\
\rowcolor{gray!20} \cellcolor{white!20} & TagFog$^\dag$ & 43.22 & 91.12 & 34.31 & 93.36 & \underline{32.62} & \underline{93.94} & \underline{37.17} & \underline{92.80} & 51.44 & 90.10 & \underline{72.91} & \underline{79.89} & \underline{45.28} & \underline{90.20} \\ 
\rowcolor{gray!20} \cellcolor{white!20}& \textbf{Ours} & \textbf{10.47} & \textbf{97.75} & 31.48 & 94.33 & \textbf{27.13} & \textbf{94.74} & \textbf{25.85} & \textbf{94.79} & 44.11 & 90.16 & \textbf{69.90} & \textbf{82.12} & \textbf{34.82} & \textbf{92.31} \\

\bottomrule

\end{tabular}%
}

\end{table*}

\begin{table}[t]
\centering
\caption{Comparison between different methods in OOD detection on CIFAR-10 and CIFAR-100 benchmarks with WideResNet28-10. All Valuesare average percentages over 6 OOD test datasets.}
\label{tab:cifar-wrn benchmark}
\small
\resizebox{1.0\linewidth}{!}{
\begin{tabular}{ccc|cc}
\toprule
\multirow{2}{*}{\textbf{Method}} & \multicolumn{2}{c}{\textbf{CIFAR-10}} & \multicolumn{2}{c}{\textbf{CIFAR-100}}\\
& \textbf{FPR95}$\downarrow$ & \textbf{AUROC}$\uparrow$ & \textbf{FPR95}$\downarrow$ & \textbf{AUROC}$\uparrow$ \\
\hline
MSP & 36.74 & 92.53 & 74.14 & 82.43 \\
ODIN & 31.71 & 88.17 & 64.66 & 84.86\\
Mahalanobis & 71.21 & 70.50  & 58.54 & 84.68 \\
Energy &  30.96 & 92.05 &  69.04 & 84.96 \\
ViM &  20.88 & 95.24 &  50.75 & 88.81 \\
DICE & 33.29 & 89.34 & 67.72 & 82.57 \\
ReAct & 26.37 & 93.91 & 67.78 & 86.15 \\
DICE+ReAct & 33.21 & 90.28 & 64.17 & 85.60  \\
FeatureNorm & 13.53 & 97.33 & 63.93 & 76.92 \\
LINe & 27.39 & 92.33 & 65.11 & 85.08  \\
\hline
\rowcolor{gray!20} CSI &  23.16 &  96.07 & 68.07 &  84.31 \\
\rowcolor{gray!20} SSD+ & 11.46 & 97.75 & \underline{45.20} & 90.09\\
\rowcolor{gray!20} VOS & 30.16 & 92.77 & 73.90 & 82.65  \\
\rowcolor{gray!20} LogitNorm & 16.25 & 96.51 & 58.42 & 86.23 \\
\rowcolor{gray!20} CIDER & 12.63 & 97.73  & 51.40 & 86.60 \\
\rowcolor{gray!20} TagFog$^\dag$ & \textbf{9.95} & \underline{97.84} & 45.56 & \underline{90.54}\\
\rowcolor{gray!20} \textbf{Ours} & \underline{10.86} & \textbf{98.07} & \textbf{33.29} & \textbf{93.03} \\

\bottomrule
\end{tabular}%
}
\end{table}

\subsection{Experimental Setup}
\noindent\textbf{Datasets}: Our method is extensively evaluated on the widely used CIFAR benchmarks~\citep{cifar} and the large-scale ImageNet~\citep{imagenet} benchmark. 
For CIFAR benchmarks, CIFAR-10 and CIFAR-100~\citep{cifar} are respectively used as in-distribution datasets, with 50,000 training images and 10,000 test images per dataset. Six OOD datasets are used during testing, including SVHN~\citep{SVHN}, LSUN-Crop~\citep{LSUN}, LSUN-Resize~\citep{LSUN}, iSUN~\citep{isun}, Textures~\citep{textures}, and Places365~\citep{zhou2017places}. For the ImageNet benchmark, ImageNet-100~\citep{NPOS} is used as the in-distribution dataset and four OOD datasets are used during testing, including Places365~\citep{zhou2017places}, Textures~\citep{textures}, iNaturalist~\citep{van2018inaturalist}, and SUN~\citep{sun}. \\

\noindent\textbf{Implementation details}: Following the common experimental setting~\citep{cider, logitnorm, NPOS}, ResNet-34~\citep{ResNet} and WideResNet28-10~\citep{wrn} are used as the backbones on CIFAR benchmarks. ResNet-50 is used as the backbone on the large-scale ImageNet benchmark. In the pre-training stage, each training set image is randomly cropped to 32 $\times$ 32 pixels (for CIFAR benchmarks) or 224 $\times$ 224 (for the ImageNet benchmark) and then perform data augmentation on training set images through random horizontally flipping and color jitter. 
In the fine-tuning stage, we remove color jitter to enable the model to identify background features more precisely.

The cross-entropy loss is used in the model pre-training stage, while the joint loss with local feature fine-tuning loss and cross-entropy loss is used in the fine-tuning stage. 
In the fine-tuning stage, we used SGD as the optimizer with the momentum as 0.9 and weight decay as 0.0001. For CIFAR benchmarks, we train the model for 30 epochs with the initial learning rate as 0.1, which then decays by a factor of 10 at epochs 15. For the ImageNet benchmark, we train the model for 10 epochs with the initial learning rate as $5\times 10^{-5}$, which then decays by a factor of 10 at epochs 5. 
For the hyperparameters, $\delta=0.1$ is applied for CIFAR-10 benchmarks and $\delta=0.01$ is applied for CIFAR-100 and ImageNet benchmarks, $\mu$ is set to be $1,2,3$ correspondingly for CIFAR-10, CIFAR-100, and ImageNet benchmarks, and $\lambda$ is set to be 1 for all benchmarks. The following section presenting the hyperparameter sensitivity experiment shows that the reasonable range of hyperparameter value selection usually leads to promising performance of our method, proving the robustness of our method with respect to hyperparameter selection. \\

\noindent\textbf{Evaluation metrics}: We measure the performance of OOD detection using the two most widely used evaluation metrics FPR95 and AUROC. FPR95 is the false positive rate when the true positive rate is 95\%, with lower FPR95 indicating better OOD detection performance and vice versa.  AUROC is the area under the receiver operating characteristic curve, with larger AUROC indicating better performance. \\

\noindent\textbf{Competitive methods for comparison}: Our method is compared with multiple competitive post-hoc OOD detection methods, including MSP~\citep{MSP}, ODIN~\citep{ODIN}, Mahalanobis~\citep{mahala}, Energy~\citep{energy}, ViM~\citep{vim}, DICE~\citep{dice}, ReAct~\citep{react}, DICE+ReAct~\citep{dice}, FeatureNorm~\citep{featurenorm}, and LINe~\citep{LINe}, as well as training-based methods, including CSI~\citep{csi}, SSD+~\citep{ssd}, VOS~\citep{VOS}, LogitNorm~\citep{logitnorm}, NPOS~\citep{NPOS}, CIDER~\citep{cider}, and TagFog~\citep{cjk1}.

\begin{table*}[t]
\centering
\caption{Comparison between different methods in OOD detection on the ImageNet Benchmark with ResNet-50. Bold numbers are superior results and underlined numbers are the 2nd best results. All values are percentages.}
\label{tab:imagenet100} 
\resizebox{\textwidth}{!}{%
\begin{tabular}{ccccccccccc}
\toprule
\multicolumn{1}{c}{\multirow{3}{*}{\textbf{Method}}} & \multicolumn{8}{c}{\textbf{OOD Datasets}} &\multicolumn{2}{c}{\multirow{2}{*}{\textbf{Average}}} \\ \cline{2-9} 
\multicolumn{1}{c}{} & \multicolumn{2}{c}{\textbf{iNaturalist}} & \multicolumn{2}{c}{\textbf{SUN}} & \multicolumn{2}{c}{\textbf{Places}} & \multicolumn{2}{c}{\textbf{Textures}}  \\
\multicolumn{1}{c}{} & \multicolumn{1}{c}{FPR95$\downarrow$} & \multicolumn{1}{c}{AUROC$\uparrow$} & \multicolumn{1}{c}{FPR95$\downarrow$} & \multicolumn{1}{c}{AUROC$\uparrow$} & \multicolumn{1}{c}{FPR95$\downarrow$} & \multicolumn{1}{c}{AUROC$\uparrow$} & \multicolumn{1}{c}{FPR95$\downarrow$} & \multicolumn{1}{c}{AUROC$\uparrow$} & \multicolumn{1}{c}{FPR95$\downarrow$} & \multicolumn{1}{c}{AUROC$\uparrow$} \\ \midrule

MSP & 69.28 & 85.84 & 70.14 & 84.20 & 69.43 & 84.29 & 64.27 & 84.09 & 68.28 & 84.60\\
ODIN & 44.22 & 92.42 & 54.71 & 88.94 & 57.52 & 88.01 & 42.87 & 89.76 & 49.83 & 89.78 \\
Mahalanobis & 96.60 & 45.22 & 98.01 & 42.55 & 97.77 & 43.87 & 38.44 & 87.88 & 82.70 & 54.88\\
Energy & 64.60 & 89.08 & 62.70  & 88.03 & 60.70  & 87.78  &51.38 & 87.89 & 59.85 & 88.20 \\
ViM  & 84.92 & 81.92 & 83.18 & 81.47 & 81.45 & 81.55 & \underline{20.05} & \underline{96.07} & 67.39 & 85.25 \\
DICE & 35.08 & 93.20 & 36.89 & 92.53  & 43.71 & 90.66 & 31.84 & 92.08 & 36.46 & 92.11\\
ReAct  & 30.60 & 94.40 & 47.55 & 89.99 & 47.21 & 89.53 & 50.89 & 87.57 & 44.06 & 90.45 \\
DICE+ReAct & \underline{26.75} & \underline{94.69} & 35.99 & 92.40 & 43.48 & 90.52 & 32.76 & 91.94 & {34.75} & 92.39 \\
FeatureNorm & 65.14 & 83.59 & 64.79 & 83.07 & 72.88 & 78.78 & 38.76 & 89.78 & 60.39 & 83.80 \\
LINe & {27.38} & {94.64} & 37.28 & 91.95 & 42.32 & 90.44 & 36.78 & 91.01 & 36.19 & 92.01 \\
\cline{1-11}

\rowcolor{gray!20} CSI & 54.25 & 87.76 & 65.13 & 82.73 & 64.68 & 84.67 & 40.27 & 90.76 & 56.08 & 86.48\\
\rowcolor{gray!20} SSD+ & 39.06 & 93.76 & 47.30 & 91.90 & 56.46 & 88.21 & \textbf{7.97} & \textbf{97.44} & 37.70 & 92.83\\
\rowcolor{gray!20} VOS & 56.40 & 88.92 & 47.50 & 90.00 & 63.20 & 87.69 & 64.30 & 85.68 & 57.85 & 88.07\\
\rowcolor{gray!20} LogitNorm & 39.73& 93.09& \underline{34.08} & \underline{93.10}& \underline{37.78}& \underline{92.53}& 42.99& 90.75&38.64 &92.37 \\
\rowcolor{gray!20} NPOS & 32.45 & 94.50 & 39.47 & 92.56 & 47.78 & 90.55 & 24.82 & 91.30 & 36.13 & 92.23\\
\rowcolor{gray!20} CIDER & 72.68 & 85.95 & 44.68 & 92.22 & 53.52 & 90.30 & {21.22} & {95.97} & 48.03 & 91.11\\
\rowcolor{gray!20} TagFog$^\dag$ & \textbf{25.58} & \textbf{95.09} & 39.21 & 91.68 & 40.75 & 91.54 & 28.46 & 93.73 & \underline{33.50} & \underline{93.01} \\
\rowcolor{gray!20} \textbf{Ours} & 34.62 & 94.23 & \textbf{25.93} & \textbf{95.05} & \textbf{36.80} & \textbf{92.77} & 29.98 & 93.85 & \textbf{31.83} & \textbf{93.97} \\

\bottomrule
\end{tabular}%
}

\end{table*}

\subsection{Evaluation on CIFAR Benchmarks}
Table~\ref{tab:cifar-resnet benchmark} summarizes the comparisons between our method and competitive OOD detection methods on CIFAR-10 and CIFAR-100 benchmarks with ResNet-34.
Our method achieved state-of-the-art performance on the CIFAR-100 benchmark, and the second best performance on the CIFAR-10 benchmark. TagFog~\citep{cjk1} is better than us on the CIFAR-10 benchmark because an additional fake OOD dataset is used for model training. 
On the CIFAR-100 benchmark, our method improves AUROC by 2.11\% and FPR95 by 10.46\% compared with the best baseline TagFog. 
Compared with VOS~\citep{VOS} and NPOS~\citep{NPOS} that synthesize fake OOD features for training, our method outperforms them by 45.71\% and 11.02\% in FPR95, respectively, which shows the effectiveness of fake OOD features we extracted and our fine-tuning strategy for OOD detection. 
In addition, our method achieves significant improvement compared to LogitNorm~\citep{logitnorm} that aims to mitigate the over-confidence issue of models, further indicating the superiority of the proposed method.
Similar results can be observed with WideResNet28-10 in Table~\ref{tab:cifar-wrn benchmark}, where our method outperforms all the compared methods on the CIFAR-100 benchmark and demonstrates superior performance on the CIFAR-10 benchmark
All the above results from CIFAR benchmarks show that our method is generally effective and has strong generalizability a different model backbones.

\subsection{Evaluation on the ImageNet Benchmark}

Table~\ref{tab:imagenet100} shows the OOD detection performance of our method and competitive baselines with the ResNet-50 backbone on the ImageNet benchmark. 
It shows that our method still achieves state-of-the-art OOD detection performance on the ImageNet benchmark. 
Compared with TagFog~\citep{cjk1} which uses an extra fake OOD image dataset, our method reduces FPR95 by 1.67\% and improves AUROC by 0.96\%.
Also, our method outperforms VOS~\citep{VOS} and NPOS~\citep{NPOS} by 26.02\% and 4.03\% in FPR95, respectively. 
This further indicates that our method not only extracts simpler fake OOD features using ID data but also can optimize the model more effectively.
Note that our method mainly achieves the best OOD detection performance on SUN and Places datasets, while both of them are natural scene datasets and include many scene images that share a similar background with images from ImageNet-100. 
Such a confusion in background is the common reason why most of the methods don't perform well on these two datasets.
Besides, our method does not perform outstandingly on the Texture dataset, with lower performance compared to ViM~\citep{vim}, SSD+~\citep{ssd}, and CIDER~\citep{cider} because the feature distribution of the Texture dataset in the feature space is significantly different from that of natural images.
ViM uses the combination of feature and logit information to perform OOD detection, while SSD+ and CIDER leverage contrastive learning to produce a more compact feature space, hence these methods have certain advantages in detecting OOD data like the Texture dataset. 
However, our method mainly shows its effectiveness when the OOD data are semantically similar to the ID data, which makes it challenging to achieve the optimal OOD detection results on the texture dataset.
In conclusion, all results support the effectiveness and superiority of our method.

\subsection{Sensitivity Study}

\begin{figure*}[!thb]
    \centering
    \subfloat[$\delta$ (CIFAR100)]{\includegraphics[width=0.32\textwidth]{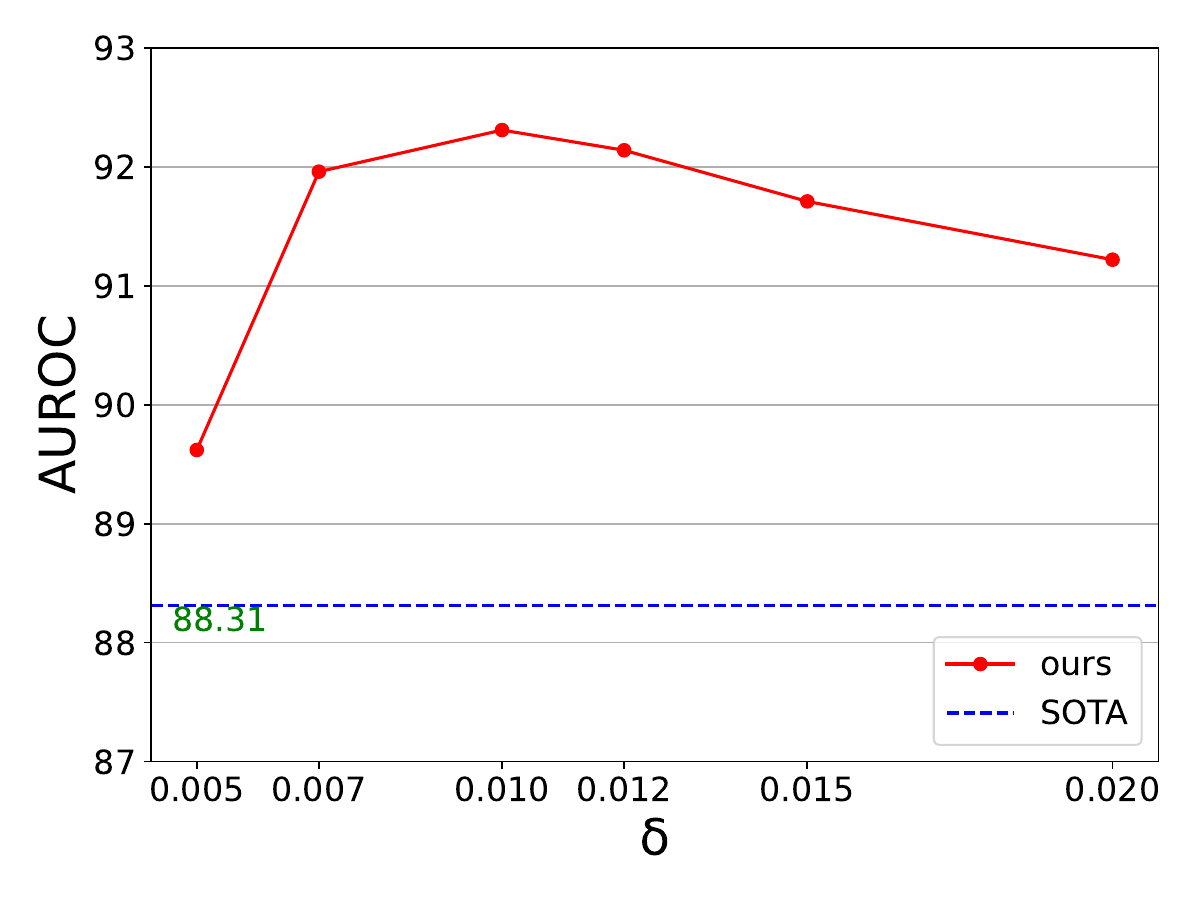}%
    \label{fig:/CIFAR-100_delta}}
    \hfil
    \subfloat[$\mu$ (CIFAR-100)]{\includegraphics[width=0.32\textwidth]{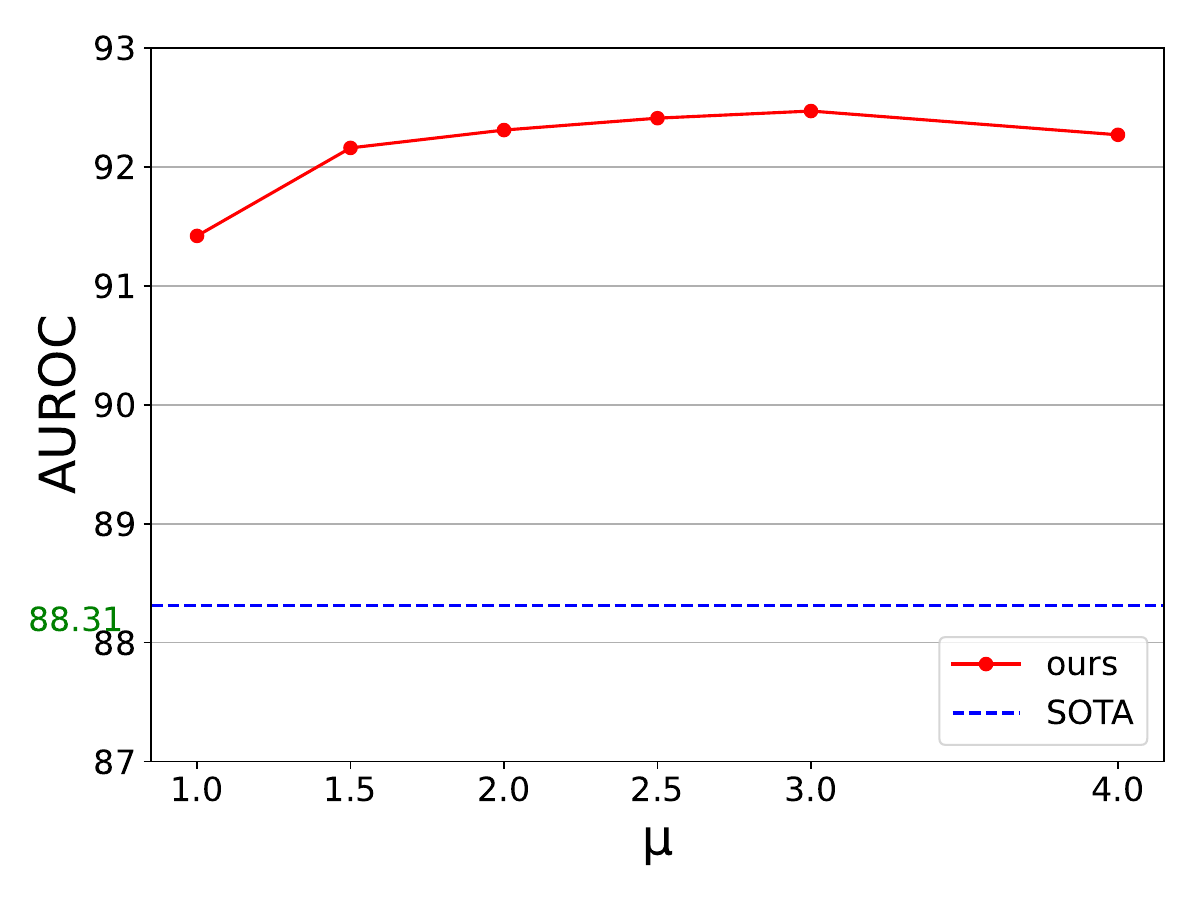}%
    \label{fig:/CIFAR-100_mu}}
    \hfil
    \subfloat[$\lambda$ (CIFAR-100)]{\includegraphics[width=0.32\textwidth]{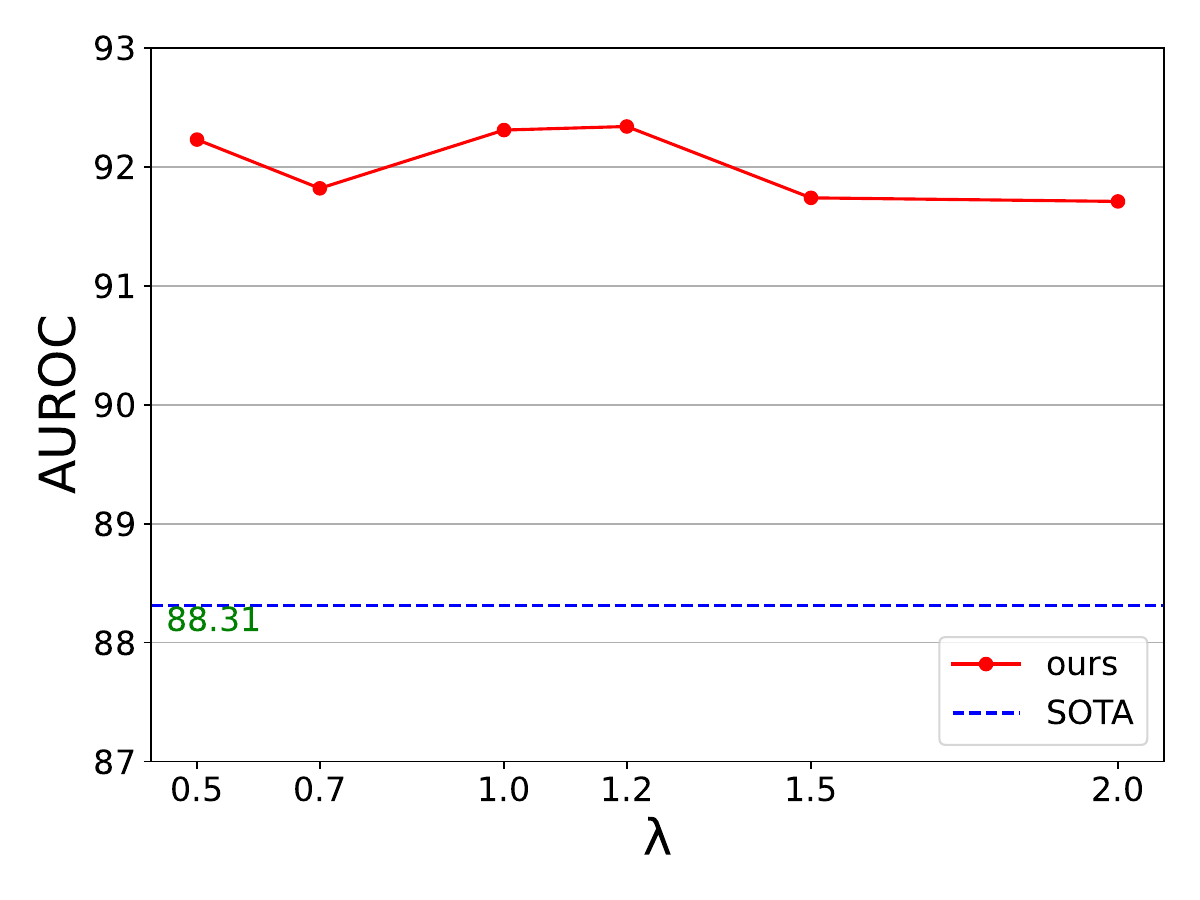}%
    \label{fig:/CIFAR-100_lambda}}
    \hfil
    \subfloat[$\delta$ (ImageNet)]{\includegraphics[width=0.32\textwidth]{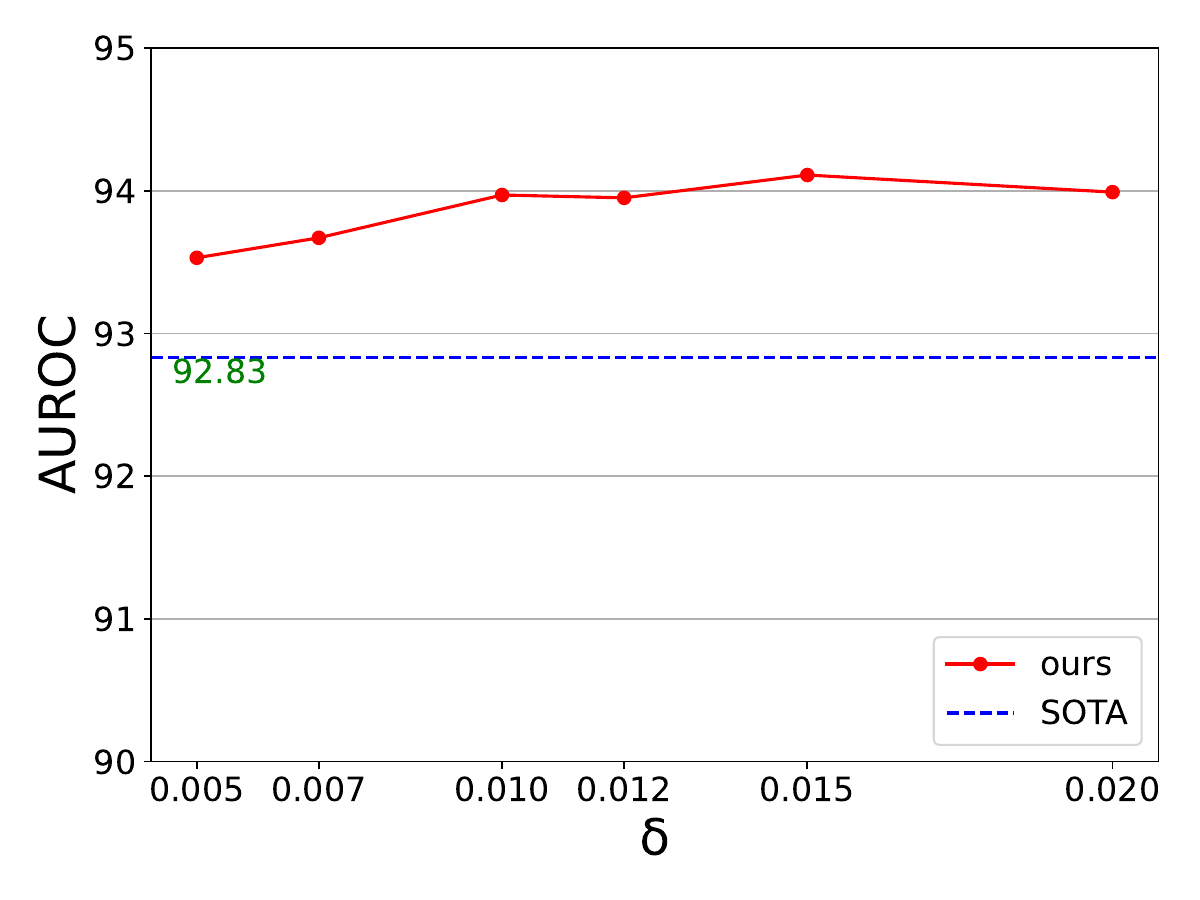}%
    \label{fig:/imagenet-100_delta}}
    \hfil
    \subfloat[$\mu$ (ImageNet)]{\includegraphics[width=0.32\textwidth]{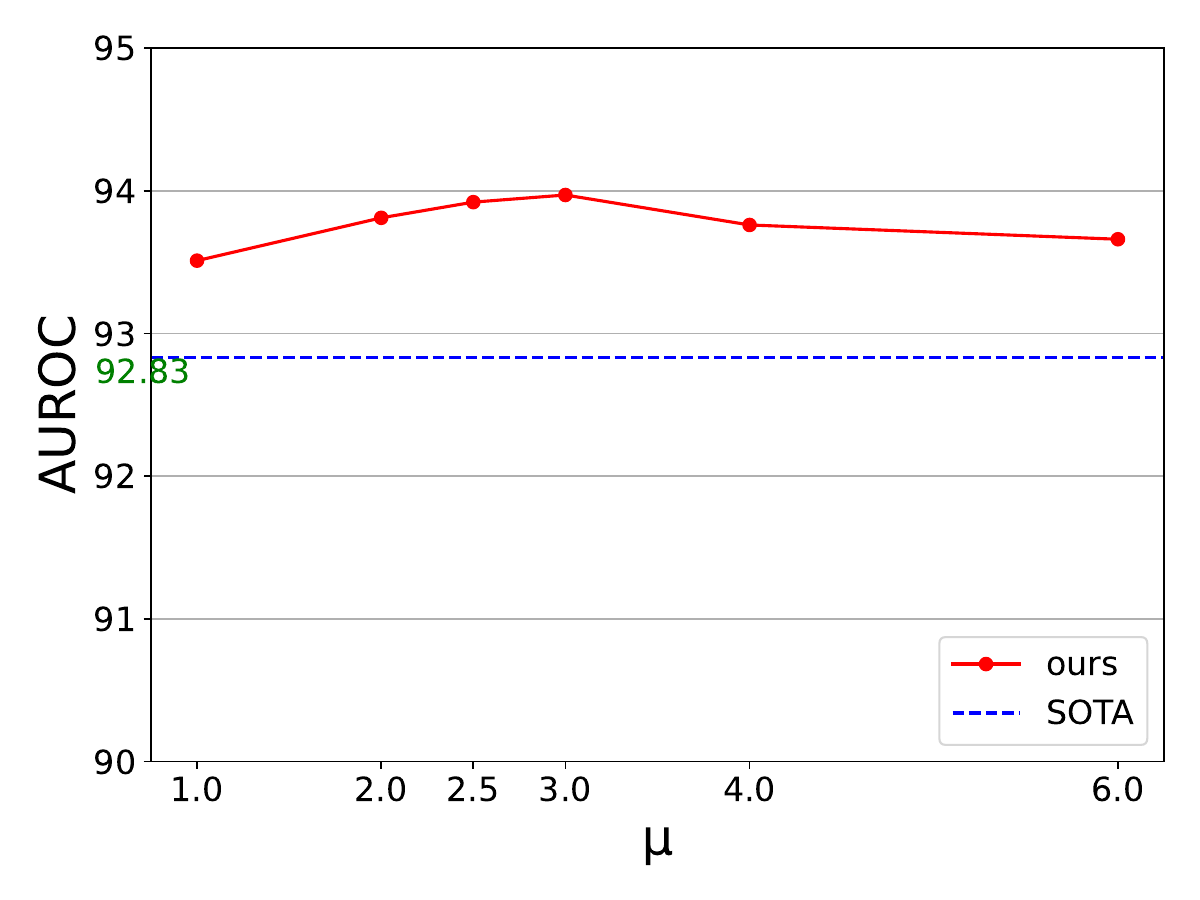}%
    \label{fig:/imagenet-100_mu}}
    \hfil
    \subfloat[$\lambda$ (ImageNet)]{\includegraphics[width=0.32\textwidth]{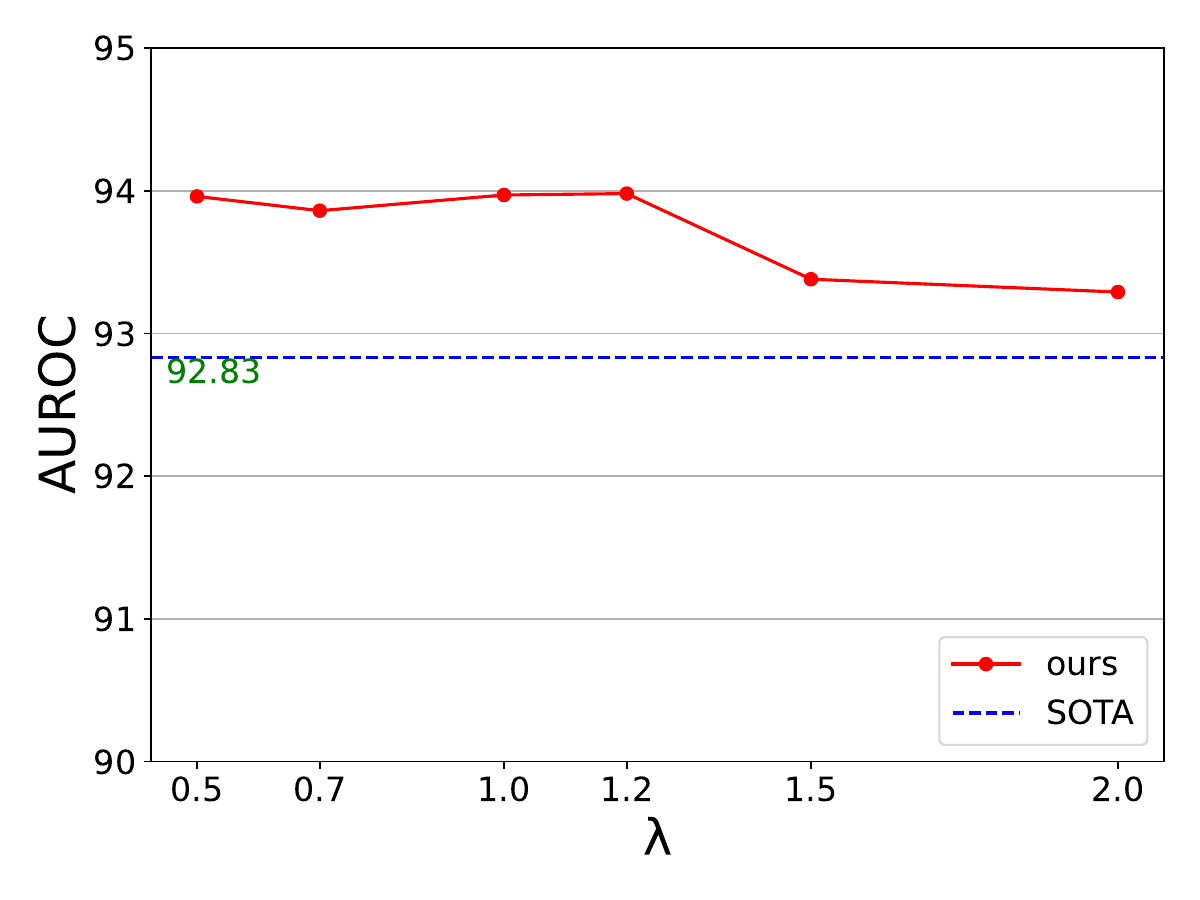}%
    \label{fig:/imagenet-100_lambda}}
    \hfil
    \caption{Sensitivity study of hyperparameters $\delta$, $\mu$, and $\lambda$ on the CIFAR-100 and ImageNet benchmarks. 
   ResNet-34 is used on CIFAR-100 benchmark, and ResNet-50 on the ImageNet benchmark.
    All values are percentages and averaged over multiple OOD datasets.  
    }
    \label{fig3:p_threshold Ablation}
\end{figure*}

Given that the selection of hyperparameters including probability threshold $\delta$, norm-margin $\mu$, and loss item weighting coefficient factor $\lambda$ will affect the overall performance, we perform a sensitivity study of the hyperparameters.
Figure~\ref{fig3:p_threshold Ablation} shows the AUROC performance of our method (red curves) on the CIFAR-100 and ImageNet benchmarks with the changing of hyperparameter. 
The dashed blue lines represent the performance of the state-of-the-art method NPOS~\citep{NPOS} (for CIFAR-100) and SSD+~\citep{ssd} (for ImageNet-100) that solely trained on the ID dataset.
As shown in Figure~\ref{fig3:p_threshold Ablation}, our method is stable and better than the strong baseline when $\delta\in[0.007,0.012]$, $\mu\in[1.5,4.0]$, and $\lambda\in[0.5,1.5]$. When $\delta$ is too small (i.e. $\delta\leq 0.005$), only a few local background features of low quality are involved for model training, leading to the small effectiveness bring by our method. 
Notably, on the ImageNet-100 benchmarks, our method still achieves superior performance even with the continuous increase of $\delta$ because there are more background regions presented in large-size images and models trained on the cross-entropy loss will pay significant attention to these background areas, which is one of the reasons for the OOD detection bottlenecks.

\subsection{Generalizability Study}

\begin{table}[!thb]
\centering
\caption{Results of applying our method to different OOD detection methods. ResNet-34 is used on CIFAR benchmarks, and ResNet-50 on the ImageNet benchmark. 
All values are averaged over multiple OOD datasets.}
\label{different method}
\resizebox{1.0\linewidth}{!}{

\begin{tabular}{ccccccc}
\toprule
\multirow{2}{*}{\textbf{Method}} & \multicolumn{2}{c}{\textbf{CIFAR-10}} & \multicolumn{2}{c}{\textbf{CIFAR-100}} & \multicolumn{2}{c}{\textbf{ImageNet}} \\
&FPR95$\downarrow$ &AUROC$\uparrow$ &FPR95$\downarrow$ &AUROC$\uparrow$ &FPR95$\downarrow$ &AUROC$\uparrow$ \\

\midrule
MSP & 40.95 & 92.09 & 78.29 & 79.25 & 68.28 & 84.60\\
MSP+Ours & \textbf{19.77} & \textbf{96.61} & \textbf{48.37} & \textbf{88.16} & \textbf{41.60} & \textbf{91.72} \\
\midrule
Energy & 26.69 & 93.17 & 69.41 & 83.64 & 59.85 & 88.20\\
Energy+Ours & \textbf{15.83} & \textbf{97.25} & \textbf{35.20} & \textbf{92.11} & \textbf{33.36 }& \textbf{93.68} \\
\midrule
ODIN & 33.06 & 89.64 & 64.01 & 83.44 & 49.83 & 89.78  \\
ODIN+Ours & \textbf{14.41} & \textbf{97.42} & \textbf{34.61} & \textbf{92.32} & \textbf{30.62} & \textbf{94.04} \\
\midrule
ReAct & 27.76 & 93.29 & 50.56 & 88.30 & 44.06 & 90.45\\
ReAct+Ours & \textbf{15.35} & \textbf{97.30} & \textbf{34.82} & \textbf{92.31} & \textbf{31.83 }& \textbf{93.97} \\
\midrule
FeatureNorm & 29.76 & 93.10 & 59.60 & 80.12 & 60.39 & 83.80\\
FeatureNorm+Ours & \textbf{20.81} & \textbf{96.25} & \textbf{41.07} & \textbf{89.57} & \textbf{46.81}& \textbf{85.66} \\
\bottomrule
\end{tabular}
}

\end{table}

The proposed method belongs to training regularization method, which means that it can be combined with multiple scoring function from different post-hoc methods to achieve better performance. Table~\ref{different method} shows the results of the combination of our method and post-hoc methods like MSP~\citep{MSP}, Energy~\citep{energy}, ODIN~\citep{ODIN}, ReAct~\citep{react} and FeatureNorm~\citep{featurenorm} on CIFAR and ImageNet benchmarks. 
The performance of the combination of these methods with our method gains a huge boost compared to the original one, which further indicates the effectiveness and wide generalizability of the proposed method. 
From the first 4 lines of Table~\ref{different method}, after fine-tuned by our method, post-hoc methods that based on the logit output can achieve promising even state-of-the-art performance (i.e. ODIN and ReAct).
This result demonstrates that our method can help solve the overconfidence issue in OOD detection, thereby improving the OOD detection performance.
Additionally, the result of our method combined with FeatureNorm indicates that the fine-tuning strategy on the local background features can effectively reduce the feature norm of real OOD data produced by the model, thus producing features with lower activation and alleviating the overconfidence issue.

\subsection{Visualization Result}

\begin{figure*}[t]
\centering
\includegraphics[width=1.0\textwidth]{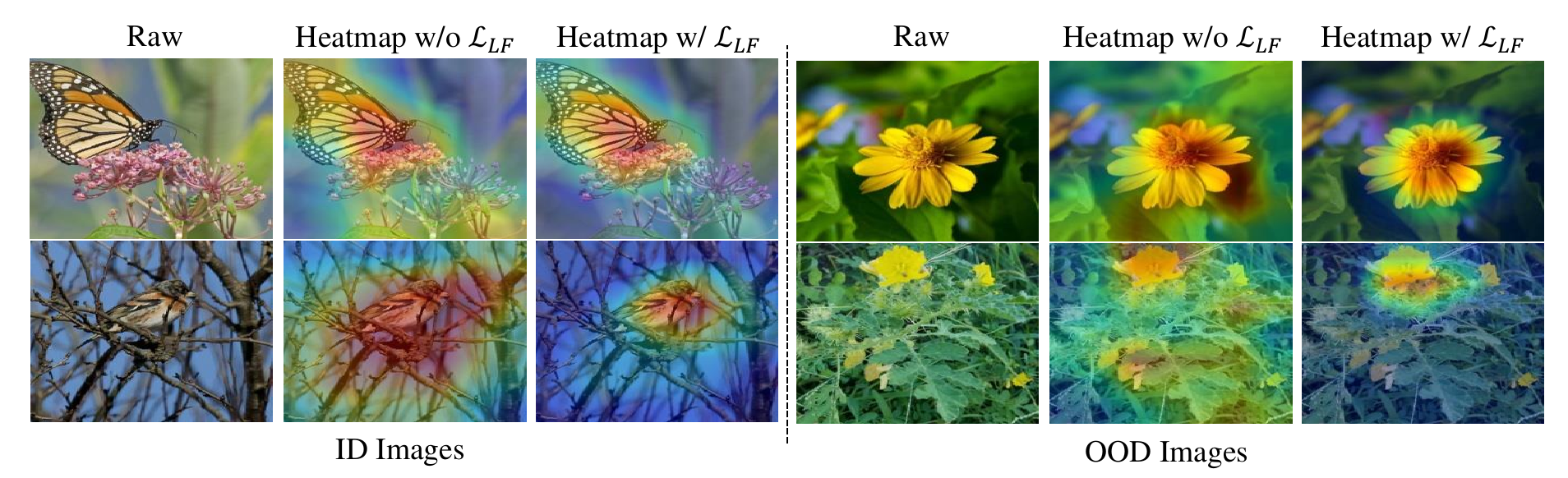} 
\caption{Heat map visualization results for ID images (ImageNet-100) and OOD images (iNaturalist, Places) under different models. The red part of the image represents higher feature activation.
Raw: raw images. Heatmap w/o $\mathcal{L}_{LF}$: visualization results under pre-trained model. Heatmap w/ $\mathcal{L}_{LF}$: visualization results under model trained by $\mathcal{L}_{CE}$ and $\mathcal{L}_{LF}$.
}
\label{fig4:visualize}
\end{figure*}

To illustrate the impacts of our method, we give the heatmap visualizations of some ID and OOD images using the pre-trained model and the model fine-tuned by our method, respectively. 
As shown in Figure~\ref{fig4:visualize}, the pre-trained model generates relatively high activations for some background regions of ID images (i.e. flowers and branches around the butterfly and the bird), which indicates that the model has paid a certain degree of attention to the background in images. 
As a result, it will also generate relatively high activations for the similar background regions in OOD images, thus making it difficult to recognize these OOD data.
However, our method can not only result in a dramatic decrease of inappropriate attention to background regions and effectively increase the attention to the category object foreground regions (i.e. Birds and Butterflies) during learning training, but also allow the model to pay less attention to the similar background regions in OOD data images. 
Although our method only extracts information from the local feature of background regions from ID data, such information still contains certain feature information of OOD data and can be leveraged to effectively enhance the OOD detection capability.

\section{Conclusion}
In this study, a novel fine-tuning framework for OOD detection is proposed by leveraging the local background features extracted from ID images.
Specifically, to address the issues of the model lacking OOD knowledge and making overconfident predictions for OOD data, we encourage the model to reduce the activation intensity of local background features during fine-tuning.
Extensive experiments demonstrate that our method establishes new state-of-the-art performance on multiple benchmarks when trained solely on ID datasets and is robust to the choice of hyperparameters. 
Moreover, the flexible combination of our method with different post-hoc methods suggests the high extensibility of our method. 
We expect that this study can provide inspiration for future research from the perspective of extracting OOD information from ID data for OOD detection.

\bibliographystyle{IEEEtran}
\bibliography{refs}

\vfill

\end{document}